\documentclass[conference]{IEEEtran}
\IEEEoverridecommandlockouts
\usepackage{cite}
\usepackage{amsmath,amssymb,amsfonts}
\usepackage{algorithm}
\usepackage{multirow}
\usepackage{tabularx}
\usepackage{enumerate}
\usepackage{algpseudocode}
\usepackage{graphicx}
\usepackage{textcomp}
\usepackage{xcolor}
\usepackage{makecell}
\usepackage{changepage}
\usepackage{booktabs}
\usepackage[letterpaper, margin=0.75in, bottom=0.75in]{geometry}
\usepackage{afterpage}

\usepackage{pgfplots}

\usepgfplotslibrary{external}

\newcolumntype{Y}{>{\centering\arraybackslash}X}

\def\BibTeX{{\rm B\kern-.05em{\sc i\kern-.025em b}\kern-.08em
    T\kern-.1667em\lower.7ex\hbox{E}\kern-.125emX}}

\renewcommand{\max}{\mathrm{max}}
\newcommand{\U}{\mathcal{U}}
\newcommand{\cam}{\mathrm{cam}}
\newcommand{\outlier}{\mathrm{outlier}}
\newcommand{\noise}{\mathrm{noise}}
\newcommand{\tabl}{\mathrm{table}}

\newcommand{\normal}{\mathcal{N}}

\newcommand\numberthis{\addtocounter{equation}{1}\tag{\theequation}}

\newcommand\blfootnote[1]{%
  \begingroup
  \renewcommand\thefootnote{}\footnote{#1}%
  \addtocounter{footnote}{-1}%
  \endgroup
}

\begin{document}

\title{\vspace*{0.2in}Bayes3D: fast learning and inference in structured generative models of 3D objects and scenes}
\author{\IEEEauthorblockN{Nishad Gothoskar\textsuperscript{*1}, Matin Ghavami\textsuperscript{*1},
Eric Li\textsuperscript{*1}
\\
Aidan Curtis\textsuperscript{1}, Michael Noseworthy\textsuperscript{1},
Karen Chung\textsuperscript{1}, Brian Patton\textsuperscript{2}\\
William T. Freeman\textsuperscript{1}, Joshua B. Tenenbaum\textsuperscript{1}, Mirko Klukas\textsuperscript{1}, Vikash K. Mansinghka\textsuperscript{1} 
}
\thanks{$^{1}$Massachusetts Institute of Technology $^{2}$Google}
}

\maketitle
\blfootnote{\textsuperscript{*}Equal Contribution}
\begin{abstract}
Robots cannot yet match humans’ ability to rapidly learn the shapes of novel 3D objects and recognize them robustly despite clutter and occlusion. We present Bayes3D, an uncertainty-aware perception system for structured 3D scenes, that reports accurate posterior uncertainty over 3D object shape, pose, and scene composition in the presence of clutter and occlusion. Bayes3D delivers these capabilities via a novel hierarchical Bayesian model for 3D scenes and a GPU-accelerated coarse-to-fine sequential Monte Carlo algorithm. Quantitative experiments show that Bayes3D can learn 3D models of novel objects from just a handful of views, recognizing them more robustly and with orders of magnitude less training data than neural baselines, and tracking 3D objects faster than real time on a single GPU. We also demonstrate that Bayes3D learns complex 3D object models and accurately infers 3D scene composition when used on a Panda robot in a tabletop scenario.
\end{abstract}


\section{Introduction}

There is a widespread need in robotics for 3D scene perception systems that can learn objects from just a handful of frames of data and robustly recognize them in clutter and high occlusion. Although neural network models have made significant progress, training them from scratch typically requires large datasets and compute budgets, and they can struggle to perform robustly. This paper introduces Bayes3D, a novel 3D scene perception system that learns 3D object models from just 1-5 frames in realtime, and robustly parses 3D scenes containing these objects, reporting coherent uncertainty about scene composition and geometry.

Bayes3D is based on GPU-accelerated sequential Monte Carlo inference in a probabilistic program that generates 3D objects and scenes. During inference, objects are detected and sequentially incorporated into a 3D scene graph model that supports massively parallel, low-resoluti on rendering and robust, hierarchical Bayesian scoring against real depth images. Object poses are inferred via coarse-to-fine enumeration, enabling scoring of large numbers of high-resolution poses at relatively low computational cost. Unlike previous probabilistic programming approaches to 3D scene perception, these innovations in model robustness and inference performance enable Bayes3D to work on challenging real-world, real-time tabletop robotics problems.

Experiments on a Panda robot show that Bayes3D can acquire complex 3D object models and robustly recognize them in practice. Qualitative demonstrations show that Bayes3D reports coherent uncertainty in challenging settings with heavy occlusion. This paper also presents quantitative benchmarks of Bayes3D's data efficiency, when tested both in-distribution and out-of-distribution, showing orders-of-magnitude improvement over convolutional neural network baselines.

\begin{figure*}
    \centering
\includegraphics[width=\textwidth]{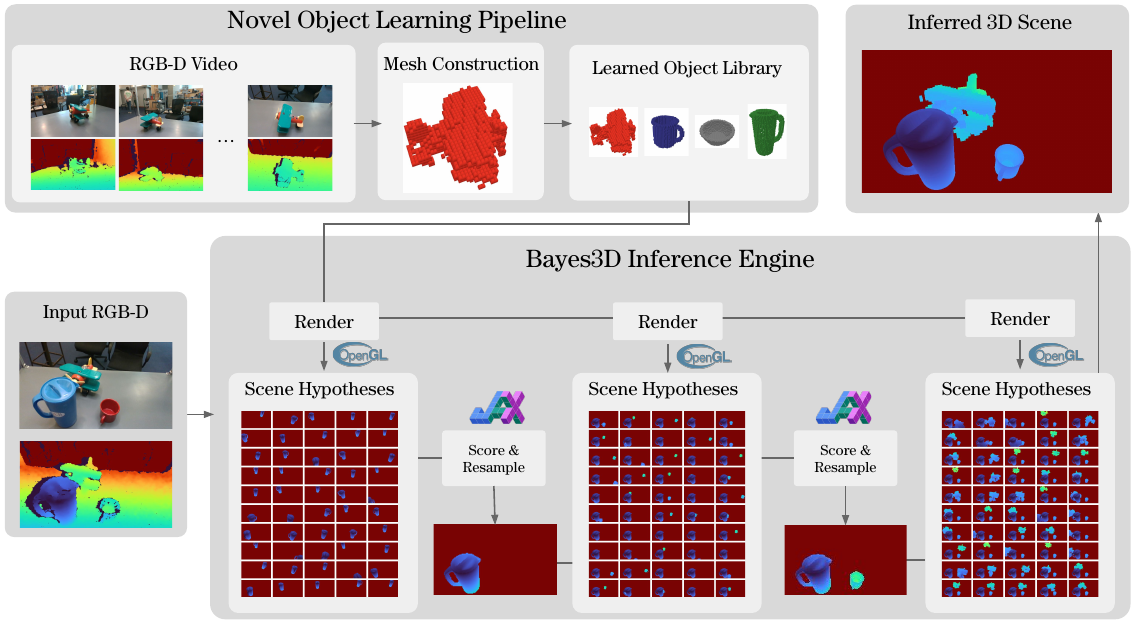}
    \caption{\textbf{Bayes3D Pipeline.} Bayes3D is an uncertainty-aware 3D scene perception system that can learn to recognize and localize novel objects from just a handful of views. To learn a novel object, we capture 5-10 RGB-D images (with a calibrated camera) from different viewpoints and overlay the resulting point clouds to get a complete point cloud representation of the object. From this, we construct a voxel mesh by discretizing the cloud at a specified resolution and placing voxels at each point. Now, given an input RGB-D image, we iteratively parse objects into a scene graph. At each step, we use a coarse-to-fine procedure to recognize and localize objects. The procedure first coarsely enumerates many scene hypotheses, evaluates their likelihood, then samples from the resulting approximate posterior. The procedure is then repeated according hyperparameters defined by a fixed schedule. We apply this coarse-to-fine procedure iteratively to infer object poses, eventually reconstructing the full scene.}
    \label{fig:modeling-inference-overview}
\end{figure*}

\section{Related Work}

\noindent \textbf{Deep learning.} Many popular approaches to 3D scene perception and pose estimation use deep learning \cite{tremblay2018deep, xiang2017posecnn, labbe2020cosypose,labbe2022megapose,li2018deepim,lipson2022coupled,manhardt2018deep}, often fusing RGB and depth data \cite{wang2019densefusion, tian2020robust} and incorporating probabilistic losses \cite{desingh2019efficient, chen2019grip, glover2012monte,deng2021poserbpf}. Outside of robotics, large neural networks for subproblems such as feature extraction \cite{oquab2023dinov2} and segmentation are also increasingly popular. These approaches typically require significant training data and compute and can struggle to robustly detect heavily occluded objects while failing to report coherent uncertainty over 3D scene composition \cite{gothoskar20213dp3}.

\noindent \textbf{Inverse graphics.} Bayes3D falls within the analysis-by-synthesis paradigm \cite{yuille2006vision, kersten2004object, lee2003hierarchical, knill1996introduction}, in which 3D perception is formulated as approximate inversion of a rendering process. Recently, differentiable formulations such relying on NeRFs \cite{mildenhall2020and, hoffman2023probnerf, yu2021pixelnerf} and 3D Gaussians \cite{kerbl20233d, keselman2022approximate} have received significant attention, but unlike Bayes3D, these typically require large numbers of images to train and do not parse scenes into 3D scene graphs \cite{tewari2023diffusion}, . 

Probabilistic programming formulations have been also developed for broad classes of 2D and 3D scenes \cite{mansinghka2013approximate, kulkarni2015picture, gothoskar20213dp3, zhou20233d}. Although these approaches are more data efficient than neural approaches, they have relied on slow, generic MCMC inference, unlike Bayes3D. The algorithms in Bayes3D are more similar to template matching approaches \cite{avetisyan2019end, avetisyan2019scan2cad} that efficiently evaluate geometric models against depth data. But unlike classical template matching, Bayes3D leverages a hierarchical Bayesian model for robust scoring even when data is noisy and a sequential Monte Carlo algorithm for efficiently inferring the composition of multi-object scenes. To the best of our knowledge, Bayes3D is the first inverse graphics system to deliver both real-time learning and high-quality approximate Bayesian inference, without relying on neural networks or generic MCMC.

\section{Methodology}

We cast pose estimation as approximate Bayesian posterior inference in a
generative model of scene depth images. A high-level overview of our approach is depicted
in Figure \ref{fig:modeling-inference-overview}. We implement our generative
model and approximate inference algorithm in a JAX-based \cite{jax2018github, 50530} GPU
implementation of Gen \cite{cusumano2019gen}, a probabilistic programming language \cite{van2018introductiont} with programmable inference \cite{mansinghka2018probabilistic}.

\subsection{Generative model}
\begin{algorithm}[tbph]
\caption{Bayes3D's generative model}
\label{alg:model}
\begin{algorithmic}[1]
\algrenewcommand\algorithmicindent{1.0em}%
\Require $n$ the number of objects appearing in the scene
\Require $O_\tabl$ a mesh model for the table
\Require $\mathbf{O}$ library of learned object voxel models
\Require $I$ camera intrinsics
\Require $\sigma_\text{max}$ prior parameter for noise model
\Procedure{SceneModel}{}
\State $G \gets \Call{FloatingSceneGraphNode}{\mathbf{1}_{SE(3)}, O_\tabl}$
\For{$i = 1, \ldots, n$}
  \State $o_i \sim \U\{\mathbf{O}\}$
  \State $c_i \sim \U\{1, \ldots, 6\}$
  \State $\Delta\phi_i \sim \Call{RelativePosePrior}{O_\tabl, o_i, c_i}$
  \State $\Call{AddChild}{G, o_i, c_i, \Delta\phi_i}$
\EndFor
\State $\theta_\cam \sim \Call{CameraPosePrior}{}$
\State $y \gets \Call{DepthRenderer}{G, \phi_\cam, I}$
\State $p_\outlier \sim \U(0, 1)$
\State $\sigma_\noise \sim \U (0, \sigma_{\text{max}})$ 
\State $\widetilde{y} \sim \Call{DepthImageLikelihood}{y, p_\outlier, \sigma_\noise, I}$
\EndProcedure
\end{algorithmic}
\end{algorithm}

Our generative model is given in algorithm \ref{alg:model}. Below we describe
different parts of the model in detail.
\begin{enumerate}[(i)]
\item \textbf{Scene prior (lines 2-8)}: We use a structured scene graph as our latent representation of scenes. For simplicity, we assume that objects are not stacked (i.e. the only node in the scene graph with non-zero out-degree is the root node representing the table) and that all objects are in contact with the table. This assumption can be relaxed with minimal modifications to the system as in \cite{gothoskar20213dp3}.  We assume that types and poses of objects in the scene are independent. We assume a uniform prior on the type of objects that can appear in the scene (line 4) and model contact relationships through the bounding boxes of the objects. For each object in the scene, we assume a uniform prior on which face of its bounding box is in contact with the table (line 5). Given the object type and contact face, three parameters completely determine the pose of the object relative to the table: the horizontal and vertical offset of the object with respect to the table, denoted $\Delta x$ and $\Delta y$, and a counter-clockwise rotation angle along the normal vector of the table denoted $\Delta\theta$. For simplicity of notation, we let $\Delta \phi := (\Delta x, \Delta y, \Delta \theta)$ in algorithm \ref{alg:model}. We also assume a uniform prior on these parameters for their valid range, that is:
\begin{align*}
    &p(o_i, c_i, \Delta\phi_i) = \frac{1}{\mathbf{O}}\times\frac{1}{6}\times\frac{1}{O_\tabl.\mathrm{width} - o_i.c_i.\mathrm{width}} \\
&\phantom{mmmmmmmmm.} \times \frac{1}{O_\tabl.\mathrm{height} - o_i.c_i.\mathrm{height}} \\
&\phantom{mmmmmmmmm.} \times \frac{1}{2\pi}.
\end{align*}
\item \textbf{Camera pose prior (line 9)}:
We assume a simple prior, on the pose of the camera frame. We assume that the eye of the camera looks directly at the origin of the world frame and that the camera's ``up" direction agrees with the positive $z$-axis of the world frame. The distance of the camera from the origin of the world frame and the azimuth and altitude angles of the camera pose are assumed to have uniform priors over fixed intervals.

\begin{figure*}[ht!]
    \centering
    \includegraphics[width=\textwidth]{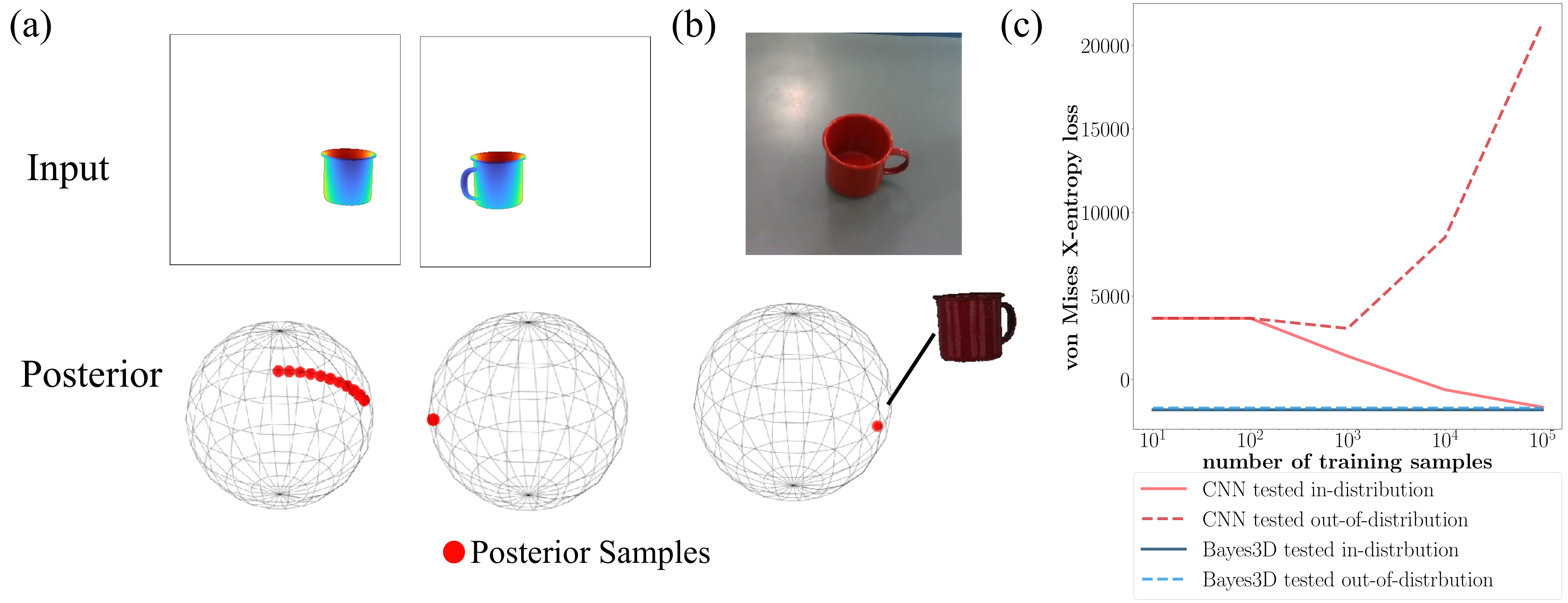}
    \caption{\textbf{6D Pose Uncertainty.} Bayes3D can infer posterior distributions over an object's 6D pose. (a) Examples from the synthetic dataset on which we evaluate and the corresponding posterior distributions inferred by Bayes3D. (b) The same procedure works on real RGB-D images. (c) Our quantitative evaluations show that Bayes3D outperforms a neural network even with substantially less training data. The neural network is also unable to generalize to out-of-distribution samples.}
    \label{fig:mug-qualitative}
\end{figure*}
\item \textbf{Depth image likelihood (lines 10-13)}:
The purpose of the likelihood is for us to be able to score a hypothesized latent scene against an observed depth image. To generate an observed image from the sampled scene and camera pose, we first use an OpenGL depth buffer to obtain a ``ground truth" depth image $y$ (line 10). We then convert this depth image into a point cloud $C$ in the camera frame. We obtain our observed depth image by creating another point cloud $\widetilde{C}$ of the same size as $C$ and converting $\widetilde{C}$ to the observed depth image $\widetilde{y}$. To obtain the $i$-th point in $\widetilde{q}_i \in \widetilde{C}$, we first flip a biased coin with probability $p_\outlier$ to determine if $\tilde{q}_i$ will be used to generate an inlier or outlier observation. Depending on whether we decide the point is an inlier or outlier, we act differently.
\begin{description}
    \item[Inlier:] If the point is decided to be an inlier, we first sample a point $q \in C$ uniformly at random, and then add independent Gaussian noise with mean 0 and variance $\sigma_\noise$ to each coordinate of $q$ to obtain $\widetilde{q}_i$.
    
    \item [Outlier:] If the point is decided to be an outlier, we sample $\widetilde{q}_i$ uniformly at random from the volume of the visible scene.
\end{description}
To ensure that our observation model can produce informative scores for a large range of images, we put priors on the noise parameters $p_\outlier$ and $\sigma_\noise$ (lines 11, 12). These priors are non-informative and exist so that, during inference, we can score hypotheses under different noise assumptions.

Putting all these together, the likelihood density is given by
\begin{align*}
&p(p_\outlier, \sigma_\noise, \tilde{y} | o_{1:n}, c_{1:n}, \Delta\phi_{1:n}) = 
    \sigma_\noise/\sigma_\max \\
& \phantom{..} \times \prod_{i=1}^{|y|}\left(
    \frac{p_\outlier}{V} + \frac{1 - p_\outlier}{|y|}\sum_{j=1}^{|y|} \normal(\tilde{y}_i; y_j, \sigma_\noise)
    \right)
\label{eq:likelihood}
\numberthis
\end{align*}
where $V$ denotes the volume of the visible scene.
\end{enumerate}


\subsection{Approximate Inference}

We use a sequential Monte Carlo (SMC) sampler \cite{del2006sequential, chopin2020introduction} with a probabilistic
program proposal \cite{lew2023smcp3} for approximate posterior inference against
Bayes3D's generative model. Using the notation of Algorithm \ref{alg:model},
our SMC sampler targets $n$ intermediate distributions. The $k$-th intermediate
distribution $p_k$ is Bayes3D's posterior given an observed depth image $\tilde y$,
but the number of objects depicted in $\tilde y$ is assumed to be $k$. More precisely,
\begin{align*}
&p_k(o_{1:k}, c_{1:k}, \Delta\phi_{1:k}, p_\outlier, \sigma_\noise| \tilde{y}) = 
\left(\prod_{i=1}^k p(o_i, c_i, \Delta\phi_i)\right) \\
&\phantom{mmm} 
p(p_\outlier, \sigma_\noise, \tilde{y}|o_{1:k}, c_{1:k}, \Delta\phi_{1:k}) / p(\tilde{y})
\end{align*}
That is,
in the first step, the SMC sampler ``explains" the observed depth image using a single
object and a large number of observed pixels are labeled as outliers. In the
second step, an additional object is added and fewer points are explained as
outliers. Eventually, the final SMC intermediate distribution is the true
posterior, explaining the scene with $n$ objects. Note that this inference
strategy starts by inferring $p_\outlier$ and $\sigma_\noise$ to be very high
and eventually lowers these estimates, thereby relying on these variables' priors on these
variables.


The proposal kernels used in Bayes3D's SMC algorithm are probabilistic programs that rely on coarse-to-fine search of the support of the posterior. At the $i$-th stage of the SMC sampler, the proposal kernel needs to propose values for variables $c_i$, $o_i$, $\Delta\phi_i$, $p_\outlier$, and $\sigma_\noise$ given 
\begin{enumerate}
    \item the observation depth image $\widetilde{y}$,
    \item the parameters $c_{1:i-1}$, $o_{1:i-1}$, $\Delta\phi_{1:i-1}$, and
    \item the estimates $(p_\outlier)_{i-1}$ and $(\sigma_\noise)_{i-1}$ of the noise parameters at stage $i - 1$
\end{enumerate}
The proposal uses the following ingredients to produce a sample:
\begin{itemize}
    \item \textbf{coarse-to-fine schedule:} Letting the support of the parameters of interest be denoted by $\mathcal{X}$, a coarse-to-fine schedule is a finite sequence of partitions $\{\Pi_t\}$ of $\mathcal{X}$ such that $\Pi_0 = \{\mathcal{X}\}$ and each element of $\Pi_t$ can be obtained by a union of a subset of elements of $\Pi_{t + 1}$. The number of partitions in the schedule indicate the number of stages in the coarse-to-fine search.
    \item \textbf{scoring strategy:} A scoring strategy for the partition $\Pi_t$ is a mapping $S_t: \Pi_t \to \mathbb{R}^+$ which assigns a positive score to each element of $\Pi_t$. In our case, these scoring functions evaluate the unnormalized SMC target at a fixed set of points within each partition cell and compute a weighted sum of the resulting values. We also truncate the sum in \eqref{eq:likelihood} for each observed pixel to a window of nearby pixel. This operation can be viewed as convolving the output of the depth renderer (line 10 of algorithm \ref{alg:model}) with a convolutional filter before applying the likelihood. This truncation introduces a small amount of bias in the importance weights of the SMC sampler but significantly improves our run time.
\end{itemize}
Given these ingredients, the proposal kernel keeps track of a subset of $A \subseteq \mathcal{X}$ from which it will sample its proposal values for the variables of interest. Initially, $A = \mathcal{X}$, indicating that the proposal can return samples anywhere in the support of the variables of interest. At stage $t$ of the coarse-to-fine search, the proposal kernel subdivides $A$ according to its schedule $\Pi_t$ and then scores each subdivided region using the scoring strategy $S_t$. The region of interest $A$ in the stage $t+1$ of coarse-to-fine is then sampled from this subdivision with probability proportional to its score. At the end, a uniform sample is generated at the final value of $A$ to be returned as the proposed values for the variables of interest. Note that our assumptions on $\Pi$ imply that this proposal distribution has a tractable density which can be used to calculate importance weights in the SMC sampler. For the rest of this section, we denote this density by $q$.

On a GPU, the scores can be calculated in parallel giving rise to a performant sampler, capable of parsing scenes in real-time. Table \ref{tab:tracking} illustrates this fact, showing various run times for camera pose calibration using such a coarse-to-fine proposal for stochastic search. 

After the proposal samples are generated, the importance weights of the particles need to be updated. The update formulas are the usual SMC updates, but since most prior terms are uniform, the weights can be slightly simplified. Equation \eqref{eq:smc-weight-update-1} gives the weight of each particle at the first stage of the SMC sampler, and \eqref{eq:smc-weight-update} shows how to obtain the weight of each particle in the $i$-th stage from the weight in the $i-1$-th stage. As is typically done in SMC samplers, we optionally resample particles based on their weights at the end of each stage.
\begin{align*}
&W_1 = 
\frac{p_1(o_1, c_1, \Delta\phi_1, p_\outlier, \sigma_\noise, \tilde{y})}
     {q(o_1, c_1, \Delta\phi_1, p_\outlier, \sigma_\noise; \tilde{y})}
\numberthis
\label{eq:smc-weight-update-1}
\end{align*}
\scalebox{0.95}{\parbox{.5\linewidth}{%
\begin{align*}
&W_i= 
\frac{W_{i-1}p_i(o_{1:i}, c_{1:i}, \Delta\phi_{1:i}, (p_\outlier)_i,
      (\sigma_\noise)_i, \tilde{y})}
     {p_{i-1}(o_{1:i-1}, c_{1:i-1}, \Delta\phi_{1:i-1}, (p_\outlier)_{i-1},
      (\sigma_\noise)_{i-1}, \tilde{y})} \\
&\phantom{.}\times
\frac{1}
     {q(o_i, c_i, \Delta\phi_i, (p_\outlier)_i, (\sigma_\noise)_i; 
        o_{1:i-1}, c_{1:i-1}, \Delta\phi_{1:i-1}, \tilde{y})}
\numberthis
\label{eq:smc-weight-update}
\end{align*}
}}

\begin{figure*}
    \centering
    \includegraphics[width=0.98\textwidth]{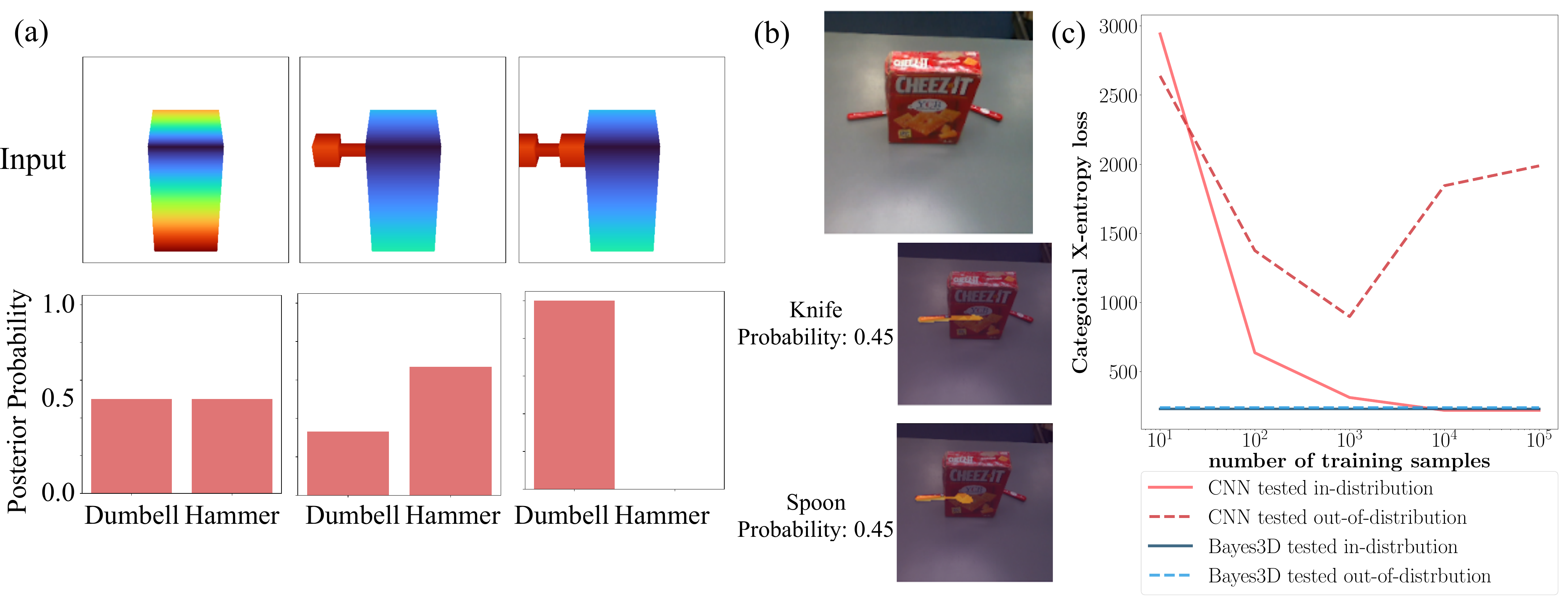}
    \caption{\textbf{Object Type Uncertainty.} Bayes3D can infer posterior distributions over an object's type. (a) Examples from the synthetic dataset on which we evaluate and the corresponding posterior over object type. (b) The same procedure works on real RGB-D images. (c) The quantitative evaluation demonstrates that Bayes3D can outperform neural architecture with substantially less data and is more robust to out-of-distribution inputs.}
    \label{fig:fork-knife-example}
\end{figure*}


\begin{figure}[ht]
    \centering
    \includegraphics[width=0.98\columnwidth]{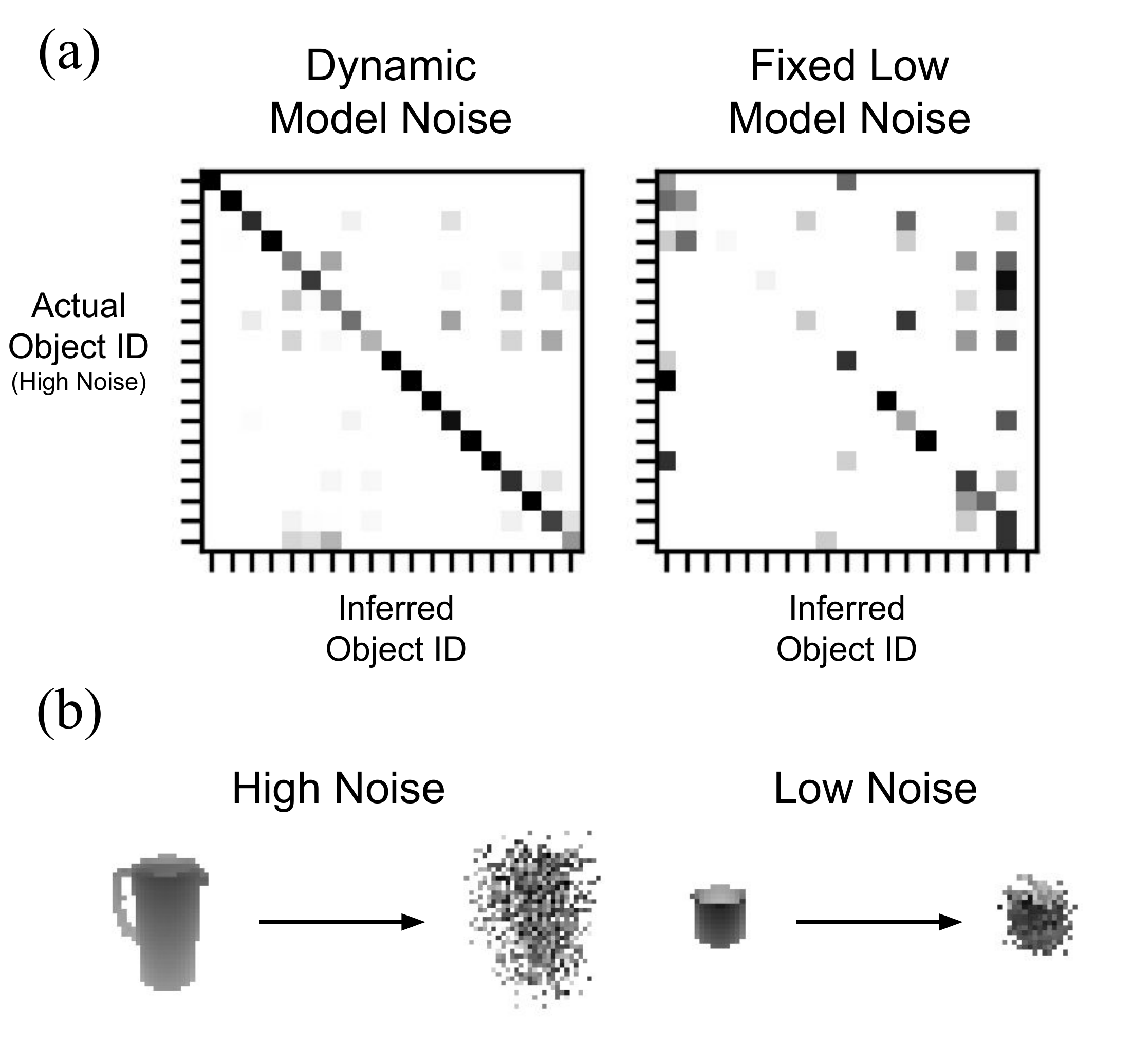}
    \caption{\textbf{Ablation Study.} A) Confusion matrices for the hierarchical model with priors over noise parameters (left) and the ablated model with fixed low noise parameters (right). The $i$'th row shows the (approximated) posterior over object identity conditioned on object $i$ (dark corresponds to high probability).
    B) Two examples of noise corrupted input depth images with high (left) and low (right) noise.}
    \label{fig:hb}
\end{figure}

\section{Experiments}

This section presents quantitative evaluations of Bayes3D's pose inferences, object class inferences, hierarchical Bayesian parameter estimates, and real-time 3D object tracking performance. We also include the results of using Bayes3D as the perception system on a real robot.

\begin{figure*}
    \centering
    \includegraphics[width=\textwidth]{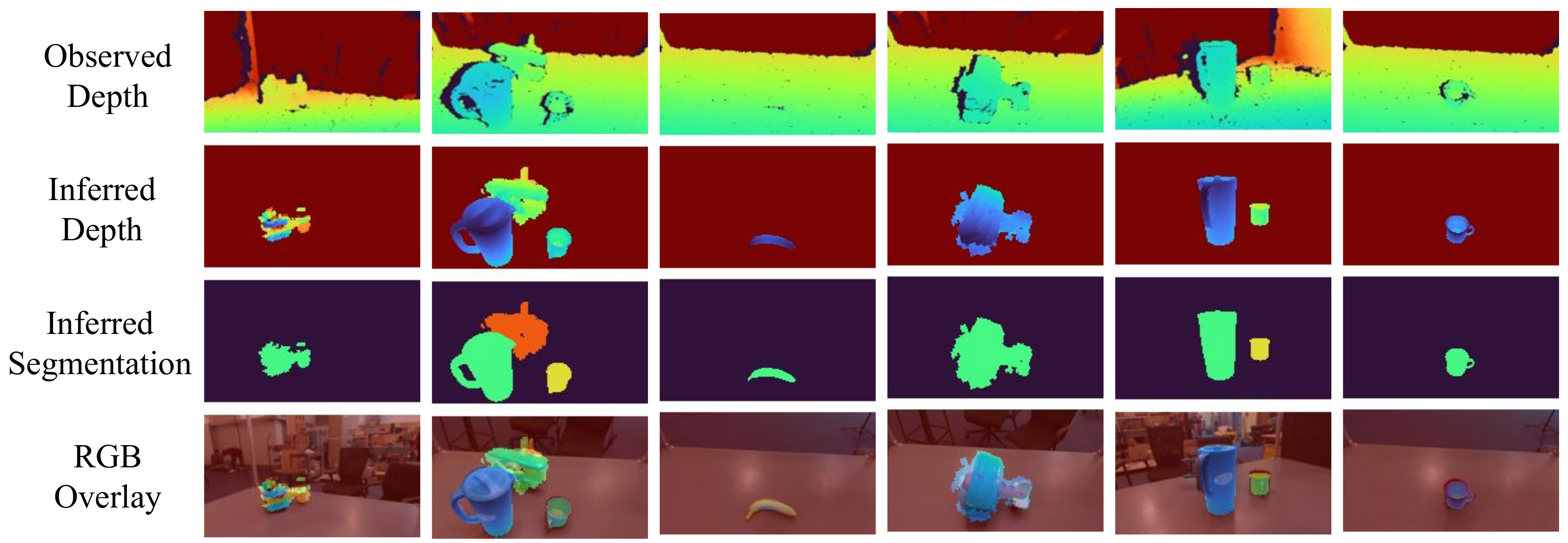}
    \caption{\textbf{Real-world scenes.} We apply Bayes3D to real-world scenes. It works on both single-object and multi-object scenes even when object models were learned independently for each object. For each scene, we show the observed depth image on which we condition our probabilistic model, the inferred depth reconstruction and segmentation mask after Bayes3D inference, and the inferred depth overlaid on the RGB input. Our real-world tests show that our system can robustly infer difficult poses and occluded objects.}
    \label{fig:real}
\end{figure*}

\subsection{6D Pose Uncertainty}


First, we compare Bayes3D's pose
uncertainty estimates with a neural pose estimator. For simplicity, we restrict
this microbenchmark to scenes with a single known object -- a YCB mug\cite{calli2015benchmarking}.
Due to self-occlusion, there will be uncertainty in the mug's pose whenever the
handle is not visible as in Figure \ref{fig:mug-qualitative}. We train a CNN \cite{lecun1989backpropagation} on synthetic
images of the mug at a fixed point on the table. The training images are labeled with the true angular component relative pose of the mug. The CNN outputs the location and concentration parameters of a von Mises distribution approximating the posterior distribution of the angular component of the mug's relative pose.
The training objective is the cross entropy loss from the true posterior on the pose to the variational
approximation predicted by the CNN.

To compare the performance of Bayes3D against the neural baseline we use Bayes3D to obtain a von Mises variational approximation. We sample 1000 particles from Bayes3D's SMC sampler and use the angular component of the relative pose of the mug to form maximum likelihood estimates of the parameters of a von Mises distribution. We then evaluate the same cross-entropy loss on the obtained variational approximation. In Figure \ref{fig:mug-qualitative} we have plotted the loss curves for Bayes3D and the CNN for both in-distribution and out-of-distribution test sets.

To achieve the same level of performance, we needed to use neural networks with roughly 7 million parameters (27.9MB), roughly 1700x more memory than the 16KB needed for Bayes3D's models.

\subsection{Object Type Uncertainty}

As a second microbenchmark we compare Bayes3D's performance on object identity inference against a neural baseline.
The dataset for this microbenchmark consists of images of scenes with an occluder
(a YCB Cheez-Its box) at a fixed pose and--with equal probability--either a dumbbell or a hammer.
Depending on the relative position of the occluder and the object, the identity of
the object might be uncertain, as shown in Figure \ref{fig:fork-knife-example}.
Our neural baseline for this microbenchmark is again a CNN, but this time with a
final softmax layer with two nodes encoding the probability of the hammer and the dumbbell. We train the network with a cross-entropy loss on synthetic data.

Figure \ref{fig:fork-knife-example} shows the results of the microbenchmark. We have evaluated the performance of the neural baselines and Bayes3D on
held-out data from the training distribution and slightly out-of-distribution data obtained by adding Gaussian noise
to the pose of the hammer or dumbbell. We can see that Bayes3D requires much less data to perform at the same level of accuracy as the neural baselines, while its quality does not deteriorate on out-of-distribution data.

To achieve this level of performance, we needed to use neural networks with roughly 3.15 million parameters (12.6MB), roughly 1050x more memory than the 16KB needed for Bayes3D's models.

\subsection{Hierarchical Bayesian Inference}

To show the necessity of having priors on the noise parameters $p_\text{outlier}$ and $\sigma_\text{noise}$, we perform an ablation study of these priors. We perform a basic classification task with the objective to infer an object's pose and identity from a single corrupted depth image; see Figure~\ref{fig:hb}B. We compare performances of a hierarchical model with uniform priors over over noise parameters ($\sigma_\text{noise}$ and $p_\text{outlier}$) and a series of ablated models with clamped sensor parameters set to a range of small to larger values. 
For each of the 19 objects from a synthetic dataset we compute the posterior probability over object identity marginalized over poses and, in case of the hierarchical model, over noise parameters. We report these posterior probabilities in form of a confusion matrix, where the $i$'th row corresponds to the (approximated) posterior conditioned on object $i$; see Figure~\ref{fig:hb}. Assuming a low noise regime, while operating in a high noise regime, results in more misclassifications; see Figure~\ref{fig:hb}A (right). 

\begin{table}[h!]
	\caption{Realtime camera pose tracking
	}
	\centering\begin{tabularx}{\columnwidth}{p{0.25in} p{0.55in}|c|cc}
		\toprule
        & & & \multicolumn{2}{c}{\textbf{Pose Accuracy}}\\
        \textbf{Object} & \textbf{Image Size} & \textbf{FPS} & \textbf{Position (cm)} & \textbf{Orientation (deg.)}\\
        \midrule
\multirow{4}{*}{Mustard}  &  25x25  &  103.207  &  0.130  &  1.524 \\
  &  50x50  &  82.129  &  0.002  &  0.841 \\
 &  100x100  &  37.944  &  0.002  &  0.736 \\
 &  200x200  &  14.507  &  0.002  &  0.739 \\
 \midrule
\multirow{4}{*}{Drill}  &  25x25  &  104.266  &  0.486  &  2.920 \\
  &  50x50  &  79.265  &  0.300  &  1.370 \\
  &  100x100  &  34.615  &  0.324  &  1.331 \\
  &  200x200  &  14.013  &  0.269  &  1.296 \\
\midrule
\multirow{4}{*}{Clamp}  &  25x25  &  104.189  &  0.650  &  3.710 \\
  &  50x50  &  84.029  &  0.392  &  1.980 \\
  &  100x100  &  37.452  &  0.240  &  1.157 \\
  &  200x200  &  14.259  &  0.254  &  1.281 \\
	\end{tabularx}
	\label{tab:tracking}
\end{table}

\subsection{Real-time 3D Tracking}

We show that Bayes3D is capable of tracking the pose of a moving camera in realtime. Just as our previous experiments involved inference of the poses of objects in the scene, the Bayes3D model supports inferring camera poses as well. We synthetically generate video sequences (120 frames) of an object panning around a single object places on the table. Using a particle filtering algorithm, we iteratively infer the camera pose at each frame, assuming the camera pose in the first frame is given. 

Table \ref{tab:tracking} shows the frame rate and accuracy of camera pose tracking for 3 different objects and 4 image resolutions. We found that at low image resolutions, Bayes3D can track at 100+ FPS and is still be fairly accurate. The speed of Bayes3D is due to (1) fast parallel rendering in OpenGL which enables rendering 2048 scene hypotheses in parallel and (2) JAX implementation of our image likelihood that allows us to score those 2048 images in parallel on the GPU.



\section{Conclusion}
In this paper, we presented an approach to uncertainty-aware 3D scene perception that can rapidly learn the shapes of novel 3D objects and then proceed to recognize and localize those objects. Our method is based on probabilistic inference in a structured generative model of 3D scenes and inference is made scalable by fast parallel coarse-to-fine SMC implemented on GPU. Our quantitative results indicate that structured uncertainty representations enable accurate, robust, and data-efficient pose inferences in real-world scenes.

\bibliographystyle{plain} 
\bibliography{refs.bib}

\begin{thebibliography}{10}

\bibitem{avetisyan2019scan2cad}
Armen Avetisyan, Manuel Dahnert, Angela Dai, Manolis Savva, Angel~X Chang, and
  Matthias Nie{\ss}ner.
\newblock Scan2cad: Learning cad model alignment in rgb-d scans.
\newblock In {\em Proceedings of the IEEE/CVF Conference on computer vision and
  pattern recognition}, pages 2614--2623, 2019.

\bibitem{avetisyan2019end}
Armen Avetisyan, Angela Dai, and Matthias Nie{\ss}ner.
\newblock End-to-end cad model retrieval and 9dof alignment in 3d scans.
\newblock In {\em Proceedings of the IEEE/CVF International Conference on
  computer vision}, pages 2551--2560, 2019.

\bibitem{jax2018github}
James Bradbury, Roy Frostig, Peter Hawkins, Matthew~James Johnson, Chris Leary,
  Dougal Maclaurin, George Necula, Adam Paszke, Jake Vander{P}las, Skye
  Wanderman-{M}ilne, and Qiao Zhang.
\newblock {JAX}: composable transformations of {P}ython+{N}um{P}y programs,
  2018.

\bibitem{calli2015benchmarking}
Berk Calli, Aaron Walsman, Arjun Singh, Siddhartha Srinivasa, Pieter Abbeel,
  and Aaron~M Dollar.
\newblock Benchmarking in manipulation research: Using the yale-cmu-berkeley
  object and model set.
\newblock {\em IEEE Robotics \& Automation Magazine}, 22(3):36--52, 2015.

\bibitem{chen2019grip}
Xiaotong Chen, Rui Chen, Zhiqiang Sui, Zhefan Ye, Yanqi Liu, R~Iris Bahar, and
  Odest~Chadwicke Jenkins.
\newblock Grip: Generative robust inference and perception for semantic robot
  manipulation in adversarial environments.
\newblock In {\em 2019 IEEE/RSJ International Conference on Intelligent Robots
  and Systems (IROS)}, pages 3988--3995. IEEE, 2019.

\bibitem{chopin2020introduction}
Nicolas Chopin, Omiros Papaspiliopoulos, et~al.
\newblock {\em An introduction to sequential Monte Carlo}, volume~4.
\newblock Springer, 2020.

\bibitem{cusumano2019gen}
Marco~F Cusumano-Towner, Feras~A Saad, Alexander~K Lew, and Vikash~K
  Mansinghka.
\newblock Gen: a general-purpose probabilistic programming system with
  programmable inference.
\newblock In {\em Proceedings of the 40th acm sigplan conference on programming
  language design and implementation}, pages 221--236, 2019.

\bibitem{del2006sequential}
Pierre Del~Moral, Arnaud Doucet, and Ajay Jasra.
\newblock Sequential monte carlo samplers.
\newblock {\em Journal of the Royal Statistical Society Series B: Statistical
  Methodology}, 68(3):411--436, 2006.

\bibitem{deng2021poserbpf}
Xinke Deng, Arsalan Mousavian, Yu~Xiang, Fei Xia, Timothy Bretl, and Dieter
  Fox.
\newblock Poserbpf: A rao--blackwellized particle filter for 6-d object pose
  tracking.
\newblock {\em IEEE Transactions on Robotics}, 37(5):1328--1342, 2021.

\bibitem{desingh2019efficient}
Karthik Desingh, Shiyang Lu, Anthony Opipari, and Odest~Chadwicke Jenkins.
\newblock Efficient nonparametric belief propagation for pose estimation and
  manipulation of articulated objects.
\newblock {\em Science Robotics}, 4(30):eaaw4523, 2019.

\bibitem{glover2012monte}
Jared Glover, Gary Bradski, and Radu~Bogdan Rusu.
\newblock Monte carlo pose estimation with quaternion kernels and the bingham
  distribution.
\newblock In {\em Robotics: science and systems}, volume~7, page~97, 2012.

\bibitem{gothoskar20213dp3}
Nishad Gothoskar, Marco Cusumano-Towner, Ben Zinberg, Matin Ghavamizadeh, Falk
  Pollok, Austin Garrett, Josh Tenenbaum, Dan Gutfreund, and Vikash Mansinghka.
\newblock 3dp3: 3d scene perception via probabilistic programming.
\newblock {\em Advances in Neural Information Processing Systems},
  34:9600--9612, 2021.

\bibitem{hoffman2023probnerf}
Matthew~D Hoffman, Tuan~Anh Le, Pavel Sountsov, Christopher Suter, Ben Lee,
  Vikash~K Mansinghka, and Rif~A Saurous.
\newblock Probnerf: Uncertainty-aware inference of 3d shapes from 2d images.
\newblock In {\em International Conference on Artificial Intelligence and
  Statistics}, pages 10425--10444. PMLR, 2023.

\bibitem{kerbl20233d}
Bernhard Kerbl, Georgios Kopanas, Thomas Leimk{\"u}hler, and George Drettakis.
\newblock 3d gaussian splatting for real-time radiance field rendering.
\newblock {\em ACM Transactions on Graphics (ToG)}, 42(4):1--14, 2023.

\bibitem{kersten2004object}
Daniel Kersten, Pascal Mamassian, and Alan Yuille.
\newblock Object perception as bayesian inference.
\newblock {\em Annu. Rev. Psychol.}, 55:271--304, 2004.

\bibitem{keselman2022approximate}
Leonid Keselman and Martial Hebert.
\newblock Approximate differentiable rendering with algebraic surfaces.
\newblock In {\em European Conference on Computer Vision}, pages 596--614.
  Springer, 2022.

\bibitem{knill1996introduction}
D.C. Knill, D.~Kersten, and A.~Yuille.
\newblock {\em Introduction}, page 1–22.
\newblock Cambridge University Press, 1996.

\bibitem{kulkarni2015picture}
Tejas~D Kulkarni, Pushmeet Kohli, Joshua~B Tenenbaum, and Vikash Mansinghka.
\newblock Picture: A probabilistic programming language for scene perception.
\newblock In {\em Proceedings of the ieee conference on computer vision and
  pattern recognition}, pages 4390--4399, 2015.

\bibitem{labbe2020cosypose}
Yann Labb{\'e}, Justin Carpentier, Mathieu Aubry, and Josef Sivic.
\newblock Cosypose: Consistent multi-view multi-object 6d pose estimation.
\newblock In {\em Computer Vision--ECCV 2020: 16th European Conference,
  Glasgow, UK, August 23--28, 2020, Proceedings, Part XVII 16}, pages 574--591.
  Springer, 2020.

\bibitem{labbe2022megapose}
Yann Labb{\'e}, Lucas Manuelli, Arsalan Mousavian, Stephen Tyree, Stan
  Birchfield, Jonathan Tremblay, Justin Carpentier, Mathieu Aubry, Dieter Fox,
  and Josef Sivic.
\newblock Megapose: 6d pose estimation of novel objects via render \& compare.
\newblock {\em arXiv preprint arXiv:2212.06870}, 2022.

\bibitem{lecun1989backpropagation}
Yann LeCun, Bernhard Boser, John~S Denker, Donnie Henderson, Richard~E Howard,
  Wayne Hubbard, and Lawrence~D Jackel.
\newblock Backpropagation applied to handwritten zip code recognition.
\newblock {\em Neural computation}, 1(4):541--551, 1989.

\bibitem{lee2003hierarchical}
Tai~Sing Lee and David Mumford.
\newblock Hierarchical bayesian inference in the visual cortex.
\newblock {\em JOSA A}, 20(7):1434--1448, 2003.

\bibitem{lew2023smcp3}
Alexander~K Lew, George Matheos, Tan Zhi-Xuan, Matin Ghavamizadeh, Nishad
  Gothoskar, Stuart Russell, and Vikash~K Mansinghka.
\newblock Smcp3: Sequential monte carlo with probabilistic program proposals.
\newblock In {\em International Conference on Artificial Intelligence and
  Statistics}, pages 7061--7088. PMLR, 2023.

\bibitem{li2018deepim}
Yi~Li, Gu~Wang, Xiangyang Ji, Yu~Xiang, and Dieter Fox.
\newblock Deepim: Deep iterative matching for 6d pose estimation.
\newblock In {\em Proceedings of the European Conference on Computer Vision
  (ECCV)}, pages 683--698, 2018.

\bibitem{lipson2022coupled}
Lahav Lipson, Zachary Teed, Ankit Goyal, and Jia Deng.
\newblock Coupled iterative refinement for 6d multi-object pose estimation.
\newblock In {\em Proceedings of the IEEE/CVF Conference on Computer Vision and
  Pattern Recognition}, pages 6728--6737, 2022.

\bibitem{manhardt2018deep}
Fabian Manhardt, Wadim Kehl, Nassir Navab, and Federico Tombari.
\newblock Deep model-based 6d pose refinement in rgb.
\newblock In {\em Proceedings of the European Conference on Computer Vision
  (ECCV)}, pages 800--815, 2018.

\bibitem{mansinghka2013approximate}
Vikash~K Mansinghka, Tejas~D Kulkarni, Yura~N Perov, and Josh Tenenbaum.
\newblock Approximate bayesian image interpretation using generative
  probabilistic graphics programs.
\newblock {\em Advances in Neural Information Processing Systems}, 26, 2013.

\bibitem{mansinghka2018probabilistic}
Vikash~K Mansinghka, Ulrich Schaechtle, Shivam Handa, Alexey Radul, Yutian
  Chen, and Martin Rinard.
\newblock Probabilistic programming with programmable inference.
\newblock In {\em Proceedings of the 39th ACM SIGPLAN Conference on Programming
  Language Design and Implementation}, pages 603--616, 2018.

\bibitem{mildenhall2020and}
Ben Mildenhall, Pratul~P Srinivasan, Matthew Tancik, Jonathan~T Barron, and
  Ravi Ramamoorthi.
\newblock and ren ng. nerf: Representing scenes as neural radiance fields for
  view synthesis, 2020.

\bibitem{oquab2023dinov2}
Maxime Oquab, Timoth{\'e}e Darcet, Th{\'e}o Moutakanni, Huy Vo, Marc
  Szafraniec, Vasil Khalidov, Pierre Fernandez, Daniel Haziza, Francisco Massa,
  Alaaeldin El-Nouby, et~al.
\newblock Dinov2: Learning robust visual features without supervision.
\newblock {\em arXiv preprint arXiv:2304.07193}, 2023.

\bibitem{50530}
Amit Sabne.
\newblock Xla : Compiling machine learning for peak performance, 2020.

\bibitem{tewari2023diffusion}
Ayush Tewari, Tianwei Yin, George Cazenavette, Semon Rezchikov, Joshua~B
  Tenenbaum, Fr{\'e}do Durand, William~T Freeman, and Vincent Sitzmann.
\newblock Diffusion with forward models: Solving stochastic inverse problems
  without direct supervision.
\newblock {\em arXiv preprint arXiv:2306.11719}, 2023.

\bibitem{tian2020robust}
Meng Tian, Liang Pan, Marcelo~H Ang, and Gim~Hee Lee.
\newblock Robust 6d object pose estimation by learning rgb-d features.
\newblock In {\em 2020 IEEE International Conference on Robotics and Automation
  (ICRA)}, pages 6218--6224. IEEE, 2020.

\bibitem{tremblay2018deep}
Jonathan Tremblay, Thang To, Balakumar Sundaralingam, Yu~Xiang, Dieter Fox, and
  Stan Birchfield.
\newblock Deep object pose estimation for semantic robotic grasping of
  household objects.
\newblock {\em arXiv preprint arXiv:1809.10790}, 2018.

\bibitem{wang2019densefusion}
Chen Wang, Danfei Xu, Yuke Zhu, Roberto Mart{\'\i}n-Mart{\'\i}n, Cewu Lu,
  Li~Fei-Fei, and Silvio Savarese.
\newblock Densefusion: 6d object pose estimation by iterative dense fusion.
\newblock In {\em Proceedings of the IEEE/CVF conference on computer vision and
  pattern recognition}, pages 3343--3352, 2019.

\bibitem{xiang2017posecnn}
Yu~Xiang, Tanner Schmidt, Venkatraman Narayanan, and Dieter Fox.
\newblock Posecnn: A convolutional neural network for 6d object pose estimation
  in cluttered scenes.
\newblock {\em arXiv preprint arXiv:1711.00199}, 2017.

\bibitem{yu2021pixelnerf}
Alex Yu, Vickie Ye, Matthew Tancik, and Angjoo Kanazawa.
\newblock pixelnerf: Neural radiance fields from one or few images.
\newblock In {\em Proceedings of the IEEE/CVF Conference on Computer Vision and
  Pattern Recognition}, pages 4578--4587, 2021.

\bibitem{yuille2006vision}
Alan Yuille and Daniel Kersten.
\newblock Vision as bayesian inference: analysis by synthesis?
\newblock {\em Trends in cognitive sciences}, 10(7):301--308, 2006.

\bibitem{zhou20233d}
Guangyao Zhou, Nishad Gothoskar, Lirui Wang, Joshua~B Tenenbaum, Dan Gutfreund,
  Miguel L{\'a}zaro-Gredilla, Dileep George, and Vikash~K Mansinghka.
\newblock 3d neural embedding likelihood for robust sim-to-real transfer in
  inverse graphics.
\newblock {\em arXiv preprint arXiv:2302.03744}, 2023.

\end{thebibliography}


\end{document}